\pdfoutput=1
\documentclass[10pt,twocolumn,letterpaper]{article}

\usepackage{iccv}
\usepackage{times}
\usepackage{epsfig}
\usepackage{graphicx}
\usepackage{amsmath}
\usepackage{amssymb}
\usepackage{booktabs}
\usepackage{multirow}
\usepackage{mathrsfs}
\usepackage{float}
\usepackage{subfig}
\usepackage{authblk}
\usepackage{array}

\usepackage[pagebackref=true,breaklinks=true,letterpaper=true,colorlinks,bookmarks=false]{hyperref}

\newcolumntype{C}[1]{>{\PreserveBackslash\centering}p{#1}}

\iccvfinalcopy 
\usepackage[capitalize]{cleveref}
\crefname{section}{Sec.}{Secs.}
\Crefname{section}{Section}{Sections}
\Crefname{table}{Table}{Tables}
\crefname{table}{Tab.}{Tabs.}

\def\iccvPaperID{5445} 

\ificcvfinal\fi

\begin{document}

\title{Crowd Counting with Sparse Annotation}

\setlength{\affilsep}{0.2em}
\author[1]{Shiwei Zhang}
\author[1]{Zhengzheng Wang}
\author[2]{Qing Liu}%
\author[1]{Fei Wang}%
\author[1]{Wei Ke\thanks{Corresponding author (wei.ke@xjtu.edu.cn).}}
\author[3]{Tong Zhang}
\affil[1]{School of Software Engineering, Xi'an Jiaotong University, Xi'an, China}
\affil[2]{School of Computer Science, Central South University, Changsha, China}
\affil[3]{School of Computer and Communication Sciences, EPFL, Switzerland }

\renewcommand*{\Authands}{, }
\maketitle
\ificcvfinal\thispagestyle{empty}\fi

\begin{abstract}
   This paper presents a new annotation method called Sparse Annotation (SA) for crowd counting, which reduces human labeling efforts by sparsely labeling individuals in an image. We argue that sparse labeling can reduce the redundancy of full annotation and capture more diverse information from distant individuals that is not fully captured by Partial Annotation methods. Besides, we propose a point-based Progressive Point Matching network (PPM) to better explore the crowd from the whole image with sparse annotation, which includes a Proposal Matching Network (PMN) and a Performance Restoration Network (PRN). The PMN generates pseudo-point samples using a basic point classifier, while the PRN refines the point classifier with the pseudo points to maximize performance. Our experimental results show that PPM outperforms previous semi-supervised crowd counting methods with the same amount of annotation by a large margin and achieves competitive performance with state-of-the-art fully-supervised methods.
\end{abstract}

\vspace{-2em}
\section{Introduction}
\label{sec:intro}

Crowd counting~\cite{zhang2016single, ma2019bayesian, li2018csrnet} is a crucial task with significant applications in public security and transportation systems. Recent advances in Convolutional Neural Networks (CNNs)~\cite{imageclass,simonyan2014very} have improved the performance of crowd-counting methods. However, most of them~\cite{song2021rethinking, ma2019bayesian, liang2022end} are fully supervised and require complete image annotations, making the annotation process challenging and time-consuming. Crowd counting annotation requires marking every head in an image with a point. This process becomes even more cumbersome when dealing with images with dense populations, where heads are small and difficult to distinguish. For example, annotating the UCF-QNRF dataset~\cite{idrees2018composition}, which contains 1535 images with an average of 815 heads per image, requires over 2000 hours.

\begin{figure}
    \centering
    \includegraphics[width=\linewidth]{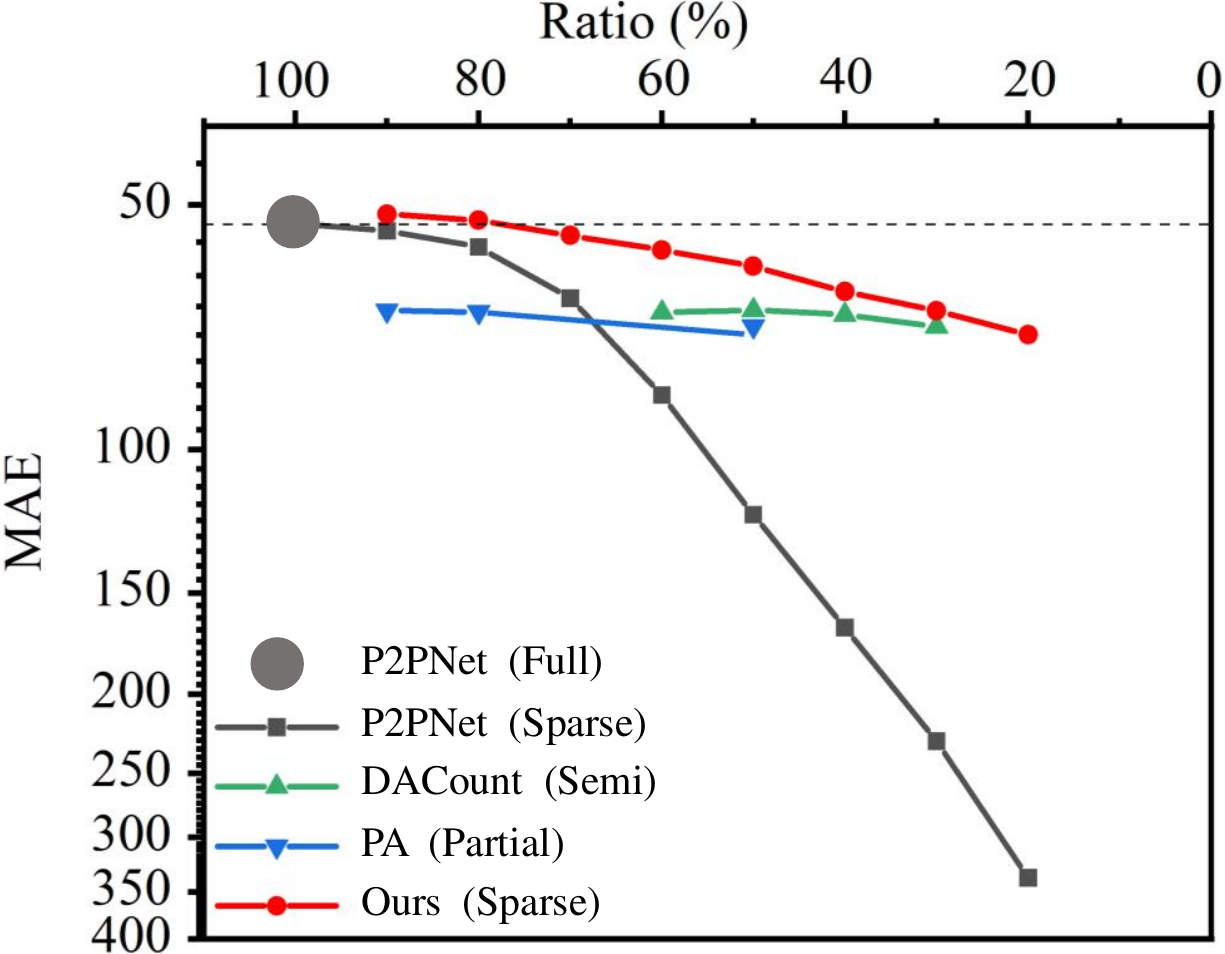}
    \vspace{-1em}
    \caption{\textbf{The MAE with different ratio }of annotation for point-based baseline P2PNet~\cite{song2021rethinking}, semi-supervised DACount~\cite{lin2022semi}, partial annotation~\cite{xu2021crowd} and our method. It illustrates that our method is closer to the upper bound of fully supervised P2PNet~\cite{song2021rethinking}.}
    \vspace{-1.5em}
    \label{fig:difper}
\end{figure}

Semi-supervised crowd counting~\cite{sindagi2020learning,liu2020semi,meng2021spatial,lin2022semi} methods can release the burden of annotation, in which a small part of the training images are fully annotated. To better utilize the information on every image, partially annotated crowd counting~\cite{xu2021crowd}
employs a patch of full annotations to train the networks. Although it incorporates more diverse information among training images with the same number of labels, it overlooks the diversity of different regions on each image. In fact, the difference between individuals in a patch is significantly less than those from distant individuals. Therefore, we propose a novel annotation setting for crowd counting in this paper named \textbf{Sparsely Annotated Crowd Counting}. We randomly and sparsely label the individuals across the whole image instead of fully annotating in a patch.
It not only captures the diversity on the image level but also incorporates the diversity at a patch level.

Most crowd counting methods are density map based ~\cite{zhang2016single,ma2019bayesian,li2018csrnet}. A gaussian kernel on every head dot is used to generate the density map. However, we cannot generate an accurate density map due to the annotation being sparse in our setting. 
To deal with this, we develop a point-based framework named Progressive Point Matching network (PPM). 
It consists of a Proposal Matching Network (PMN) and a Performance Restoration Network (PRN). 
In the PMN, a point classifier is trained to predict a confidence score for each proposal. One-to-one matching is leveraged to assign the point proposals to ground truth.
The performance of PMN is significantly influenced when the annotated ratio is low. The fewer labels we have, the easier the classifier is affected by noisy labels. To exclude the inaccurate labels and explore the benefits of the sparse labels, we incorporate a progressive proposal selection strategy to choose pseudo samples during the training procedure. Hence, the PRN, a distillation classifier that learns confident pseudo samples with high scores is proposed. Meanwhile, a confidence-guided weighted loss is applied to adaptively assign weight to the positive proposal to further alleviate the impact of noisy point samples. As shown in Fig.~\ref{fig:difper}, the performance of the baseline P2PNet~\cite{song2021rethinking} declines significantly when the annotation ratio becomes lower (gray curve), while ours remains the best across all the methods (red curve).

We conduct extensive experiments to evaluate the efficiency of the proposed PPM approach under the sparse annotation setting on the challenging datasets of ShanghaiTech\_A~\cite{zhang2016single}, ShanghaiTech\_B~\cite{zhang2016single}, UCF\_QNRF~\cite{idrees2018composition}, UCF\_CC\_50~\cite{idrees2013multi} and NWPU-Crowd~\cite{nwpu}. The results show that our method outperforms all of the previous semi-supervised crowd counting methods by a large margin. Moreover, it only needs 80\% annotations to achieve competitive performance with previous state-of-the-art fully-supervised methods and 90\% annotations to exceed them. 

In a nutshell, our main contributions can be summarized as follows:
\begin{itemize}
    \item We propose a novel crowd counting annotation setting called \textbf{Sparsely Annotated Crowd Counting} to reduce the annotation cost and capture the diversity of image level and patch level.
    \item We design a simple yet effective framework named Progressive Point Matching network (PPM), consisting of a Proposal Matching Network to train a basic point classifier supervised by sparse annotation and a Performance Restoration Network to refine the point classifier. 
    \item The proposed PPM approach on Sparsely Annotated Crowd Counting setting achieves competitive performance with the state-of-the-art fully-supervised crowd counting methods, and outperforms all the semi-supervised methods by a large margin.
\end{itemize}

\section{Related Work}
\label{sec:relate}
\subsection{Fully-supervised Crowd Counting}
The traditional crowd counting methods are mainly based on handcrafted features~\cite{chen2012feature,idrees2013multi,subburaman2012counting,chan2009bayesian,fiaschi2012learning}, which are not performed well when the scenes are extremely congested and severely occluded. 
In recent years, based on the rapid development of Deep Convolutional Neural Network, fully-supervised crowd counting gains satisfying performance.

Density map based method was firstly introduced in~\cite{lempitsky2010learning}, they used all of the point labels to produce a pseudo density map to supervise the predicted density map of network. The estimated number is obtained by summing over the predicted density map. Recently, a large amount of studies focus on density map scheme, most of them solve various challenges in crowd counting and push forward the counting performance frontiers. \cite{zhang2016single} proposed a multi-column model to regress the predicted density map, and designed an adaptive Gaussian kernels to generate pseudo density maps. In order to deal with extremely congested scenes, \cite{li2018csrnet} fused dilated kernels into crowd counting networks to estimate the count. \cite{jiang2020attention,sindagi2019ha,miao2020shallow,zhang2019relational,zhang2019attentional,lin2022boosting} introduced attention network to crowd counting to extract attentive features. In~\cite{cao2018scale,liu2019context,liu2019adcrowdnet,babu2017switching,jiang2019crowd, wei2021scene}, they used adaptive scales, trellis architecture, scale aggregation network, or deformable attention in Transformer network to solve the large scale variations in crowd counting. In ~\cite{ma2019bayesian}, the Bayesian assumption was introduced and Bayesian loss was proposed to estimate the count expectation at each annotated point. In \cite{wang2020distribution}, optimal transport was proposed to match the distributions. \cite{ma2021learning} proposed unbalanced optimal transport to estimate the discrepancy between density map and annotations. 

\begin{figure}
    \centering
    \includegraphics[width=\linewidth]{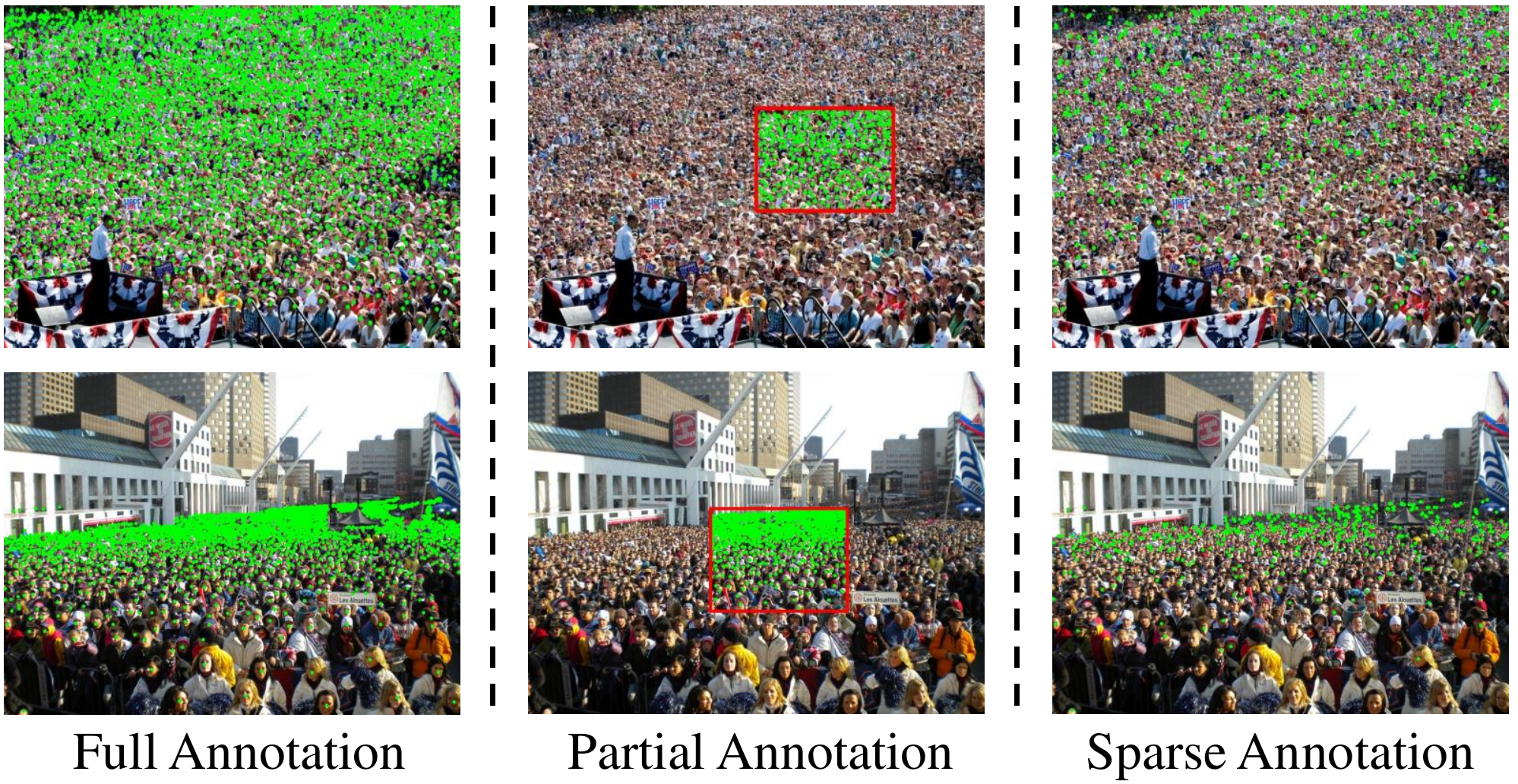}
    \vspace{-1em}
    \caption{\textbf{The three annotation settings for crowd counting} are depicted from left to right as follows: \textbf{Full Annotation}, which involves annotating all individuals in an image; \textbf{Partial Annotation}, which estimates the crowd counting using a patch of the fully annotated image; and our \textbf{Sparse Annotation}, which involves random sampling of the ground truth individuals across the image.}
    \vspace{-2em}
    \label{fig:difset}
\end{figure}

Except for density map based methods, some new solutions were also proposed. \cite{song2021rethinking} discarded density maps and proposed a purely point-based network, using Hungarian Matching to make a one-to-one matching. \cite{Liu_2021_ICCV} exploited the sample correlations by the multi-expert model. In addition, \cite{Wang_2021_ICCV} proposed Mean Count Proxies to select the best count proxy for each interval to represent its count value during inference. \cite{liang2022end} introduces transformer into crowd localization, proposing a crowd localization transformer to directly predict the localization. \cite{niu2022local} proposes a local matching point-based framework to alleviate instability. 

These fully supervised methods achieve incredible performance on various datasets. However, the complete annotation for a crowd image is time-consuming and laborious. In UCF-QNRF dataset~\cite{idrees2018composition}, it costs over 2000 human hours to annotate only 1535 images, in which the average cost per image is over one hour. 
\label{sec:method}

\begin{figure*}[htbp]
  \centering
  \includegraphics[width=\linewidth]{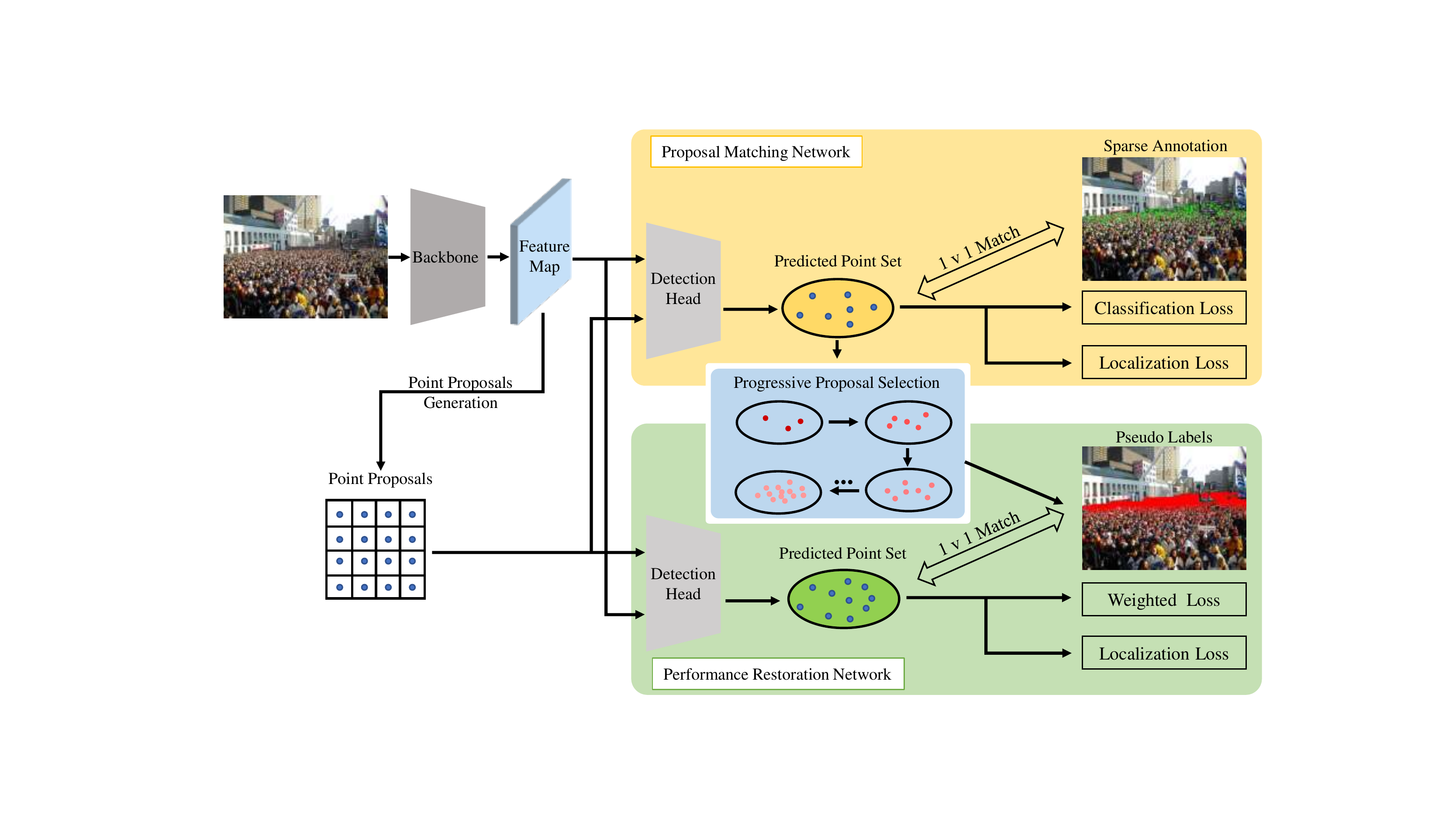}
  \vspace{-1em}
  \caption{The overall architecture of the Progressive Point Matching network. Feature map extracted by backbone is used to generate point proposals. The feature and proposals are input into two parallel networks. The Proposal Matching Network trains a basic point classifier that produces a coarse prediction and provides pseudo point samples. The Performance Restoration Network gets a distillation classifier which is supervised by the pseudo points. A Progressive Proposal Selection strategy and a confidence-guided weighted loss are used to alleviate the adverse impact of noise in pseudo samples.}
  \vspace{-1.5em}
  \label{fig:Net_Arch}
\end{figure*}

\subsection{Weakly/Semi-/Un- Supervised Crowd Counting}
In order to relieve the crowd counting annotation efforts, an increasing number of weakly/semi-/un- supervised crowd counting methods have been proposed recently.

In~\cite{liu2018leveraging}, it collected many unlabeled crowd images from the internet and leveraged them to train a learning-to-rank framework to estimate the density maps for crowd counting. Inspired by it, a soft-label sorting network is proposed to directly regress the number which is very difficult to optimize~\cite{yang2020weakly}. In~\cite{xiong2022glance}, the binary ranking of two images was used to train the network, and anchors were introduced during the inference stage to yield accurate counting numbers. Furthermore, \cite{sindagi2020learning} focused on learning to count from a limited number of annotated samples, and proposed a Gaussian Process-based iterative learning mechanism. \cite{liu2020semi} tackled the problem from the perspective of feature learning, unlabeled images are used to train a generic feature extractor. In~\cite{meng2021spatial}, the spatial uncertainty-aware teacher-student framework was proposed to deal with the noisy supervision from unlabeled data. In~\cite{lin2022semi}, a density agency and a contrastive learning loss was proposed to recognize foreground features and consolidate feature extractor. Moreover, \cite{xu2021crowd} decided to label one patch of an image rather than the entire image. Besides, \cite{babu2022completely} only need the approximate upper limit of the crowd count for the given dataset, proposing a completely self-supervised crowd counting framework. \cite{liang2022transcrowd} develop a counting method using count-level annotations but this method can not reduce annotation effort.

In this paper, we propose a solution that is significantly distinct from all of the above methods. Specifically, to capture the diversity on both the image level and patch level, Sparsely Annotated Crowd Counting(SACC) is designed. With random and sparse annotation, we propose a Progressive Point Matching network (PPM) to explore the benefits of sparse annotation. The solution outperforms all previous semi-supervised methods with the same amount of annotation and obtains competitive performance with fully-supervised methods which achieve a balance between performance and annotation cost.


\section{The Setting of Sparse Annotation}
\label{subsec:form}

We denote the training image set for crowd counting as $\mathcal{X}$ with the head point $\mathcal{P}$ and test image set $\mathcal{Y}$ with counting number $\mathcal{N}$. Given an image $\textit{I} \in \mathcal{X}$ with $N$ people, the set of head point is $P = \{p_i\}_{i=1}^N \in \mathcal{P}$, where $p_i = (x_i,y_i)$ is the coordinate of the center point for the $i$-th person. 

Traditional semi-supervised methods completely label part of images in the training set, as shown on the left of Fig.~\ref{fig:difset}. It ignores the image level difference. Partial annotation setting increases the diversity on the image level by employing a patch of full annotation, as shown in the middle of Fig.~\ref{fig:difset}. However, it overlooks the difference in the patch level. By observing that the difference between individuals in a patch is largely less than those from distant individuals, we propose a novel setting of sparse annotation to capture the diversity on both the image level and the patch level. As shown on the right of Fig.~\ref{fig:difset}, in this new setting, we randomly and sparsely annotate the individuals across the whole image. 

As the accurate $N$ cannot be obtained for the training image in our setting during annotation, we can roughly estimate the total number $\hat{N}$ 
and ask the annotator to label $N^c = ratio \times \hat{N}$ people\footnote{In this paper, we directly random select $N^c = ratio \times N$ points as sparse annotation from the completely annotated dataset.}. The $ratio$ controls the number of head points to be annotated, which is steerable to make a trade-off between performance and annotation cost.
The set of annotated head points for sparse annotation is denoted as ${P}^c \in \mathcal{P}^c$ and $\mathcal{P}^c$ is a subset of $\mathcal{P}$.  
The ultimate goal is to train a model on $\mathcal{X} \times \mathcal{P}^c$ to predict a counting number on $Y \in \mathcal{Y}$ which is close enough to the accurate counting number $N \in \mathcal{N}$.

\section{The Proposed Method}

\subsection{Overview}

To deal with sparsely annotated crowd counting, we utilize a point-based scheme
rather than generating inaccurate density map
due to the nature of sparse annotation. We propose a simple yet effective Progressive Point Matching network (PPM), as shown in Fig.~\ref{fig:Net_Arch}. The training image is passed through a VGG16 Network~\cite{simonyan2014very} to produce feature maps. Then these features are fed into a Proposal Matching Network(PMN) to train a basic point classifier generating pseudo head point samples. To exclude the inaccurate labels and explore the benefits of the sparse labels, we incorporate a Progressive Proposal Selection strategy to pick out the pseudo samples. At last, we propose a Performance Restoration Network(PRN) as a distillation classifier to learn the confident pseudo samples with high scores.

In the inference stage of PPM, the test image is forwarded into the PRN to predict the head points. The total number is generated by counting the predicted heads.


\subsection{Proposal Matching Network}
\label{subsec:PMN}

The PMN is used to produce a coarse prediction by the supervision of sparse annotation $\mathcal{P}^c$ and provides pseudo point samples for the PRN, as shown in Fig.~\ref{fig:Net_Arch}. 

\textbf{Point Proposal Generation.}
Supposing the down-sampling rate of the backbone is $s$, each location of the feature map is corresponding to a $s \times s$ patch of the input image. To densely detect the head points, $s \times s$ point anchors are used at the center of the patch. The point proposals are generated by sliding the point anchors on the feature map. We denote the set of point proposals as $R = \{r_j|j \in \{1,...,M\} \}$, where $r_j$ is a point proposal with location $(x_j,y_j)$ and $M$ is the total number of point proposals, which is equal to the size of an input image. 

\textbf{Point Detection.}
Similar to object detection, point detection is accomplished with two parallel branches for point classification and point coordinate regression. The classification branch predicts the confidence scores $\hat{c}_j$ for every point proposal. The regression branch outputs $M$ pairs of location offsets $(\Delta{x_j}, \Delta{y_j})$. The set of predicted points is denoted as $\hat{P} = \{\hat{p}_j|j\in \{1,...,M\}\}$, where $\hat{p}_j=(x_j + \Delta{x_j}, y_j + \Delta{y_j})$.  
With the confidence score and point coordinates, a one-to-one matching is conducted between the predict points $\hat{P}$ and sparse annotation ${P}^c$ with the Hungarian algorithm ~\cite{kuhn1955hungarian} of $\Omega (P^c,\hat{P},D)$. The pair-wise matching cost matrix $D$ is defined as $D(P^c,\hat{P}) = (\nu ||p_i-\hat{p}_j||_2 - \hat{c}_j)$,
where $p_i\in P^c$ and $\hat{p}_j \in \hat{P}$, $||\cdot||_2$ is Euclidean distance, and $\nu$ is a weight term. With the one-to-one matching, the $i$-th annotated point is assigned to one predicted point $\hat{p}_{\sigma(i)}$, where $\sigma(i)$ is the index of the matched point. The unmatched are labeled as negatives. The total loss of Proposal Matching Network is defined as:
\begin{equation}
\mathcal{L} = \mathcal{L}_{cls} + \lambda_1 \mathcal{L}_{loc}\;,
\label{eq:loss}
\end{equation}
where the classification loss $L_{cls}$ is a Cross Entropy Loss defined as:
\begin{equation}
    \begin{aligned}
        \mathcal{L}_{cls} = -\frac{1}{M}(\sum^{N^c}_{i=1}log\hat{c}_{\sigma(i)} + \lambda_2 \sum^M_{N^c+1}log(1-\hat{c}_{\sigma(i)}))\;,
    \end{aligned}
\end{equation}
and the location regression loss $L_{loc}$ is Euclidean distance loss defined as:
\begin{equation}
    \begin{aligned}
        \mathcal{L}_{loc} = \frac{1}{N^c} \sum^{N^c}_{i=1} ||p_i-\hat{p}_{\sigma(i)} ||^2_2\;.
    \end{aligned}
\end{equation}


\subsection{Performance Restoration Network}
\label{subsec:PRN}

PMN provides a basic point classifier and regressor, but the performance is greatly affected when the annotated ratio decreases. To exclude the inaccurate labels and explore the benefits of the sparse annotation, we incorporate a progressive proposal selection strategy to choose pseudo samples during the training
procedure as Fig.~\ref{fig:Net_Arch} shows. Meanwhile, the PRN, a distillation classifier that learns the confident pseudo samples with high scores is also proposed to refine the point classifier. The PMN and PRN are trained end-to-end. It consumes less training time than that in an alternative way.  


\textbf{Progressive Proposal Selection. }
The predicted points of the PMN are noisy as the number of proposals $M$ is much larger than the number of annotated points $N^c$. Thus, it is critical to choose reliable pseudo points. Intuitively, we choose the positive point proposals with a hard threshold which is explained in Appendix. 
However, because the score for each point proposal is changing during training, it is better to change the threshold value along with the increasing of training epochs.
Hence we propose a Progressive Proposal Selection strategy to exclude inaccurate labels and explore the benefits
of the sparse labels. Denote $r$ as a point sample with confidence score $c$, 
$0\leq W_t\leq 1$ as the weight applied to $r$ at epoch $t$ with $0\leq t\leq T$ satisfying $W_0=0$ and $W_T=1$.
With the increase of $W_t$, the distribution of training samples becomes more complex for the PRN. 
The pseudo points are selected by a progressive threshold depending on $W_t$. The label of the point is defined as:
\begin{equation}
L_{pseudo} = \left\{
    \begin{aligned}
        1, & \quad \text{if} \quad c \geq \tau_1 - \tau_2  \cdot W_t \\
        0, & \quad \text{otherwise}
    \end{aligned},
    \right.
\end{equation}
where $\tau_1$ and $\tau_2$ are constant values satisfying $0 \leq \tau_2$ \textless $\tau_1 \leq 1$ to limit the progressive threshold to a section $[\tau_1-\tau_2, \tau_1]$.

\textbf{Weighted Loss.}
The loss functions of the PRN is similar to the PMN as Eq.~(\ref{eq:loss}). A Hungarian Matching is conducted between the pseudo points and the point proposals so that the matched proposals are labeled as positive and the others are negative. The number of pseudo point labels is denoted as $N^s$. The loss of location regression is updated as: 
\begin{equation}
    \begin{aligned}
        \mathcal{L}_{loc} = \frac{1}{N^s} \sum^{N^s}_{i=1} ||p_i-\hat{p}_{\delta(i)} ||^2_2 \;.
    \end{aligned}
\end{equation}

A weighted classification loss is used in the PRN, defined as:
\begin{equation}
    \begin{aligned}
        \mathcal{L}_{cls} = -\frac{1}{M}(\sum^{N^s}_{i=1}w\log\hat{c}_{\delta(i)} + \lambda \sum^M_{N^s+1}\log(1-\hat{c}_{\delta(i)})) \;,
    \end{aligned}
    \label{eq:weightedcls}
\end{equation}
where $w$ denotes the scores of proposals outputted by the PMN. 
When the learning beginning, the pseudo labels are not accurate and a small weight $w$ is applied to alleviate the issue of over-fitting to noises. With the training epochs increasing, the prediction is more accurate and reliable, and the higher weight $w$ is applied. It brings the advantage that the positive ones are easier to be recognized accurately.

\section{Experiment}

We first introduce the experimental setting including datasets, implementation details, and evaluation metrics. Then, the effectiveness of the proposed Sparse Annotation and counting approach is discussed. At last, the performance is compared with the state-of-the-art semi-supervised, partially annotated, and fully supervised crowd counting approaches. Besides, comparison with crowd localization methods is discussed in Appendix. 

\subsection{Experimental setting}

\textbf{Datasets.} Our experiments are conducted on five publicly available datasets: ShanghaiTech\_A(SHA)~\cite{zhang2016single}, ShanghaiTech\_B(SHB)~\cite{zhang2016single}, UCF\_QNRF(QNRF)~\cite{idrees2018composition}, UCF\_CC\_50(UCF50)~\cite{idrees2013multi} and NWPU-Crowd(NWPU)~\cite{nwpu}. Among them, SHA is used for ablation study.





\textbf{Implementation details.} We use the VGG-16 model pre-trained on ImageNet as our backbone. The number of anchor points is set as 8 for the QNRF dataset and 4 for the others. Following~\cite{song2021rethinking}, the weight term $\nu$ is set as 5e-2, $\lambda$, ${\lambda_1}$ and${\lambda_2}$ are set as 0.4, 0.05 and 0.4. $\tau_1$ and $\tau_2$ are set as 0.6 and 0.4 empirically. We use Adam optimizer to optimize the network with a fixed learning rate 2e-4. The batch size is 8. 

We initially adopt random scaling which is selected from [0.7, 1.3]. Then we randomly crop patches from a resized image. For SHA, SHB, UCF50, and NWPU, the patch size is set to $128 \times 128$. For QNRF, the patch size is $256\times 256$. In the end, a random horizontal flipping with a probability of 0.5 is adopted. For QNRF and NWPU, we keep the length of the longer side smaller than 1408 and 2048 respectively to keep the original aspect ratio.

\textbf{Evaluation metrics.} We follow the commonly used metrics in crowed counting, using Mean Absolute Error (MAE) and Mean Squared Error (MSE) for evaluation. 
 
\subsection{Ablation Study}

\textbf{Comparison with the Baseline.}
We conduct experiments on different annotation ratios to verify the effectiveness of our method on the SHA dataset. The pure point-based method, \textit{i.e.}, P2PNet~\cite{song2021rethinking}, is taken as our baseline, as shown in Table~\ref{tab:difper}. With complete annotation, the baseline miraculously achieves MAE of 52.7 and MSE of 85.1, which is an impressive performance of fully supervised crowd counting. However, the MAE and MSE drastically deteriorate along with the decrease of the annotation ratio, especially when the ratio is set to 50\%. 
The reduction of annotation makes the baseline under-fitting. In sharp contrast, our methods trained on different ratios show a big advantage over the baseline. When the annotation ratio is 90\%, the MAE and MSE are miraculously better than the fully-supervised baseline. Following the decrement in the annotation ratio, the MAE and MSE of our approach increase slightly from 51.3 to 72.2 and from 84.3 to 134.7, respectively. 
Besides, 80\% annotations are enough for our method to surpass the performance of the fully-supervised baseline, which saves 20\% annotation labor. The proposed setting of sparse Annotation and approach is steerable to achieve a trade-off between the crowd counting performance and the annotation labor.

\begin{table}[t]
\centering
\begin{tabular}{lcccc}
\toprule
\multicolumn{1}{c}{\multirow{2}{*}{Ratio}} & \multicolumn{2}{c}{Baseline}   & \multicolumn{2}{c}{Ours}      \\
\multicolumn{1}{c}{}                       & MAE           & MSE            & MAE           & MSE           \\
\midrule
100\%                                      & 52.7          & 85.1           & --            & --            \\
90\%                                       & 53.8          & 90.5           & 51.3          & 84.3          \\
80\%                                       & 56.3          & 95.1           & 52.2          & 84.9          \\
70\%                                       & 65.1          & 112.4          & 54.5          & 91.7          \\
60\%                                       & 85.7 & 153.7 & 57.3 & 94.6 \\
50\%                                       & 120.2         & 207.9          & 59.5          & 103.9         \\
40\%                                       & 165.5         & 252.2          & 63.9          & 111.2         \\
30\%                                       & 228.4         & 316.1          & 67.5          & 121.2         \\
20\%                                       & 336.3         & 438.4          & 72.2          & 134.7       \\
\bottomrule
\end{tabular}
\vspace{-1em}
\caption{The performance comparison between the baseline and our approach with a series of annotation ratios. The MAE and MSE of the baseline are drastically degraded along with the decrease of the annotation ratio, while the performance degradation of our approach is slight.}
\label{tab:difper}
 \vspace{-0.6em}
\end{table}


\begin{table}
  \centering
  \begin{tabular}{lccc}
    \toprule
    & Weighted Loss & MAE & MSE\\
    \midrule
    Baseline w. 60\%   &   &85.7 &153.7\\
    \midrule
    Hard Threshold  & & 59.9    & 103.7\\
    Hard Threshold  & \checkmark &  58.3   &  101.3 \\
    \midrule
    PPS &  &58.4  &98.5   \\
    PPS & \checkmark &\textbf{57.3}  &\textbf{94.6}   \\
    \bottomrule
  \end{tabular}
  \vspace{-1em}
  \caption{Ablation study of Weighted Loss and PPS.}
  \vspace{-1em}
  \label{tab:ab}
\end{table}

\textbf{The effectiveness of PPS and Weighted Loss.}
As shown in Table~\ref{tab:ab}, the experiments study the effectiveness of the Progressive Proposal Selection (PPS) strategy and Weighted Loss for the Performance Restoration Network. In these experiments, the annotation ratio is set to a fixed value of 60\%. 

A simple hard threshold largely improves the baseline method with 60\% annotations. The MAE decreases from 85.7 to 59.9 and the MSE decreases from 153.7 to 103.7. These verify that the Performance Restoration Network is efficient. As the threshold varies with different datasets, we utilize the Progressive Proposal Selection strategy, in which an exponential function is used to control the change rate of parameter $W_t$. The MAE and MSE are reduced by 1.5 points and 5.2 points, respectively. When the weighted loss is used, the MAE and MSE are further reduced by 1.6 points and 2.4 points, respectively. The MAE and MSE have further decreased to 57.3 and 94.6 with 1.71\% and 3.95\% relative improvements when both PPS and Weighted Loss are adopted.

\begin{table}[t]
  \centering
  \begin{tabular}{llcccc}
  \toprule
\multirow{2}{*}{Disturbance} & Ratio      & \multicolumn{2}{c}{80\%} & \multicolumn{2}{c}{60\%} \\
                             & range      & MAE         & MSE        & MAE         & MSE        \\
\midrule
                    & $\pm$ 0              &             52.2&84.9            &             57.3&94.6            \\
$N(0,3)$            & $\pm$ 5\%            &             52.6&86.5            &             57.7&95.3            \\
$N(0,5)$            & $\pm$ 7\%            &             52.7&86.7            &             58.3&96.7            \\
$N(0,11)$           & $\pm$ 10\%           &             52.8&88.1            &             58.6&99.7             \\
$N(0,25)$           & $\pm$ 15\%           &             53.0&89.9            &             58.7&99.2  \\
\bottomrule
\end{tabular}
  \vspace{-1em}
  \caption{The impact of annotation disturbance on SHA.}
  \vspace{-0.6em}
  \label{tab:disturb}
\end{table}

\begin{table}[t]
  \centering
  \begin{tabular}{lccc}
    \toprule
    Method & Ratio & MAE &MSE\\
    \midrule
    LPMN~\cite{niu2022local} & 100\%   &52.1 & 86.4  \\
    \hline
    LPMN + PPM(ours) & 90\%   &\textbf{51.7} & \textbf{85.5}  \\
    LPMN + PPM(ours) &80\%   &53.0  &88.1\\
    \bottomrule
  \end{tabular}
  \vspace{-1em}
  \caption{LPMN~\cite{niu2022local} is adopted as PMN on SHA.}
  \vspace{-1em}
  \label{tab:p2p}
\end{table}


\begin{table*}[]
\begin{tabular}{lccccccccccc}
\toprule
\multirow{2}{*}{Method} & \multirow{2}{*}{Venue} & \multirow{2}{*}{Type} & \multirow{2}{*}{Ratio} & \multicolumn{2}{c}{SHA}       & \multicolumn{2}{c}{SHB}      & \multicolumn{2}{c}{UCF50}       & \multicolumn{2}{c}{QNRF}      \\
\cline{5-12} 
                        &                       &                       &                        & MAE           & MSE           & MAE          & MSE           & MAE            & MSE            & MAE           & MSE           \\
\midrule
MCNN~\cite{zhang2016single}                    & CVPR2016              & FS                    & 100\%                  & 110.2         & 173.2         & 26.4         & 41.3          & 377.6          & 509.1          & 277           & 426           \\
CP-CNN~\cite{sindagi2017generating}                  & ICCV2017              & FS                    & 100\%                  & 73.6          & 106.4         & 20.1         & 30.1          & 295.8          & 320.9          & --            & --            \\
CSRNet~\cite{li2018csrnet}                  & CVPR2018              & FS                    & 100\%                  & 68.2          & 115.0         & 10.6         & 10.6          & 266.1          & 397.5          & 108.2         & 181.3         \\
SANet~\cite{cao2018scale}                   & ECCV2018              & FS                    & 100\%                  & 67.0          & 104.5         & 8.4          & 13.6          & 258.4          & 334.9          & --            & --           \\
CAN~\cite{liu2019context}                     & CVPR2019              & FS                    & 100\%                  & 61.3          & 100.0         & 7.8          & 12.2          & 212.2          & \underline{243.7} & 181.3         & 183.0         \\
BL~\cite{ma2019bayesian}                      & ICCV2019              & FS                    & 100\%                  & 62.8          & 101.8         & 7.7          & 12.7          & 229.3          & 308.2          & 88.7          & 154.8        \\
ASNet~\cite{jiang2020attention}                   & CVPR2020              & FS                    & 100\%                  & 57.8            & 90.1            & --           & --            & 174.8          & 251.6          & 91.6          & 159.7       \\
LibraNet~\cite{liu2020weighing}                & ECCV2020              & FS                    & 100\%                  & 55.9            & 97.1            & 7.3           & 11.3            & 181.2          & 262.2          & 88.1          & 143.7 \\
DMCount~\cite{wang2020distribution}                 & NeurIPS2020           & FS                    & 100\%                  & 59.7          & 95.7          & 7.4          & 11.8          & 211.0          & 291.5          & 85.6          & 148.3          \\
GL~\cite{wan2021generalized}                      & CVPR2021              & FS                    & 100\%                  & 61.3          & 95.4          & 7.3          & 11.7          & --             & --             & \underline{84.3} & 147.5        \\
P2PNet~\cite{song2021rethinking}                  & ICCV2021              & FS                    & 100\%                  & 52.7          & 85.1 & \underline{6.5} & \textbf{10.5} & 172.7          & 256.2          & 85.3          & 154.5          \\
ChfL~\cite{shu2022crowd} &CVPR2022 &FS  & 100\%  &57.5 &94.3 &6.9 &11.0 &-- &-- &\textbf{80.3} &\textbf{137.6}\\
DC~\cite{xiong2022discrete}                      & ECCV2022              & FS                    & 100\%                  & 59.7          & 91.4          & 6.8           & 11.5            & --             & --             & 84.8            & \underline{142.3}             \\
\midrule
PPM(ours)                     &    2023       &PA                                 & 90\%                   &\textbf{51.3}  &\textbf{84.3}          & \textbf{6.4}          & \textbf{10.5}            & \textbf{163.9}              &\textbf{241.9}          &84.8   &143.1           \\
PPM(ours)                    &    2023                   & PA                    & 80\%                   & \underline{52.2} & \underline{84.9}          & 6.6          & 10.7          & \underline{170.8} & 249.9          & 85.1          & 147.4        \\

\bottomrule
\end{tabular}
\vspace{-1em}
\caption{Comparison with fully-supervised (FS) approaches. With 80\% sparse annotations (SA), our method achieves the best MAE on ShanghaiTech\_A, and UCF\_CC\_50, and competitive performances on ShanghaiTech\_B and UCF\_QNRF datasets. With 90\%, the performances are further improved consistently on all datasets. The best scores are in bold and the second ones are with an underline.}
\vspace{-1em}
\label{tab:full}
\end{table*}

\textbf{Impact of annotation ratio disturbance.}
When the annotator is asked to label the head points in an image with an accurate ratio for a crowd counting dataset, there must be some disturbance so that an approximate ratio of heads is annotated. We simulate this disturbance by applying different Gaussian distributions and using the 3$\sigma$ rule to vary the range with different $\sigma$ values. 
For example, when the disturbance follows the $N(0,11)$ distribution, the required annotation ratio is 80\% while the actual ratio ranges from 70\% to 90\%. 

From Table~\ref{tab:disturb}, we can conclude that the disturbance of the annotation ratio does not have a significant impact on the performance of our model. With $80\%$ annotations, the MAE increases by about 0.4 points for the range $\pm 3\%$, decreases by 0.5 points for the range $\pm 5\%$, and decreases 0.6 points and 0.8 points respectively for the range $\pm 10\%$ and $\pm 15\%$. With a $60\%$ annotation ratio, the largest increase observed is 1.4 points, which is relatively small. These demonstrate that the proposed approach is robust when there exists a disturbance of the annotation ratio.
Consequently, when a new crowd counting dataset is annotated  under our sparsely annotated setting, an expected accurate annotation ratio is not necessary and an approximate one is sufficient. 

\begin{table}[t]
    \begin{tabular}{p{1.8cm}ccccc}
         \toprule
         \multirow{2}{*}{Method} &\multirow{2}{*}{Ratio} &\multicolumn{2}{c}{Val} &\multicolumn{2}{c}{Test}\\
         \cline{3-6}
         &  &MAE &MSE &MAE &MSE\\
         \midrule
         CSRNet~\cite{li2018csrnet}          &100\% &104.9 &433.5 &121.3 &387.8\\
         CAN~\cite{liu2019context}           &100\% &93.5  &489.9 &106.3 &386.5\\
         BL~\cite{ma2019bayesian}            &100\% &93.6  &470.4 &105.4 &454.2\\
         DMCount~\cite{wang2020distribution} &100\% &70.5  &357.6 &88.4  &388.6\\
         P2PNet~\cite{song2021rethinking}    &100\% &--    &--    &\textbf{77.4}  &362.0  \\
         \midrule
         PPM(ours)                       &80\% &\textbf{65.1} &\textbf{311.9} & 83.7 & \textbf{353.3} \\  
         \bottomrule
    \end{tabular}
    \vspace{-1em}
    \caption{Comparison with fully-supervised approaches on the NWPU-Crowd dataset.}
    \vspace{-1em}
    \label{tab:nwpu}
\end{table}

\textbf{PPM with different point classifier.} In the previous experiments, we used P2PNet~\cite{song2021rethinking} as a point classifier to train PPM. However, it is worth noting that the PPM strategy can be applied to any point-based method. To demonstrate this, we also train PMN using LPMN~\cite{niu2022local}, another point-based approach. Quantitative results show that it also achieves better results with a 90\% ratio and competitive results with an 80\% ratio, as shown in Table~\ref{tab:p2p}.

\subsection{Comparison with Fully-Supervised Methods}
We extensively compare our method with the  fully-supervised methods on SHA, SHB, UCF50, and QNRF, as shown in Table~\ref{tab:full}.
With only 80\% annotations, we consistently reach the fully supervised baseline, \textit{i.e.}, P2PNet ~\cite{song2021rethinking} on all datasets. Compared with other approaches, our method achieves state-of-the-art performances on SHA with an MAE of 52.2 and UCF50 with an MAE of 170.8, and with competitive performances on SHB and QNRF. If the annotation rate is 90\%, the MAE and MSE are further improved, and the state-of-the-art performance on SHB is also achieved with an MAE of 6.4.

The results on the large-scale NWPU~\cite{nwpu} dataset are illustrated in Table.~\ref{tab:nwpu}. The proposed PPM achieves state-of-the-art performance with MAE of 65.1 and MSE of 311.9 on the validation set with 80\% annotations and comparable performance on the test set. These demonstrate that our PPM is robust in various crowd counting scenes.

\begin{table}[t]
\centering
\begin{tabular}{p{2.0cm}p{0.55cm}p{0.55cm}p{0.55cm}p{0.55cm}p{0.55cm}p{0.55cm}}
\toprule
\multirow{2}{*}{Method} & \multicolumn{2}{c}{50\%}       & \multicolumn{2}{c}{40\%}       & \multicolumn{2}{c}{30\%}       \\
\cline{2-7} 
    & MAE & MSE   & MAE & MSE & MAE & MSE        \\
\midrule
MT~\cite{DBLP:conf/nips/TarvainenV17}                                          & 88.2          & 151.1          &  --             &  --              &  --             &   --             \\
L2R~\cite{liu2018leveraging}                                        &  86.5          & 148.2          &   --            &  --              &   --            &   --             \\
SUA~\cite{meng2021spatial}                                        &   68.5          & 121.9          &    --           &  --              &    --           &   --             \\
DACount$^\dag$~\cite{lin2022semi}                                      & 67.3          & 113.2          & 68.2          & 116            & 70.6          & \textbf{117.8} \\
Ours                                        &  \textbf{59.5} & \textbf{103.8} & \textbf{63.9} & \textbf{111.2} & \textbf{67.5} & 121.2     \\
\bottomrule
\end{tabular}
\vspace{-1em}
\caption{Comparison with the semi-supervised crowd counting methods. $^\dag$ indicates the results reproduced by open-source code.}
\vspace{-1em}
\label{tab:semi}
\end{table}

\subsection{Comparison with Semi-Supervised Methods}

The setting of crowd counting with Sparse Annotation is a special case of semi-supervised manner. The typical semi-supervised setting is with complete annotation on part of the training images, while ours is with Sparse Annotation on each image of the training data. 

\begin{figure*}[t]
  \centering
  \includegraphics[width=0.99\linewidth]{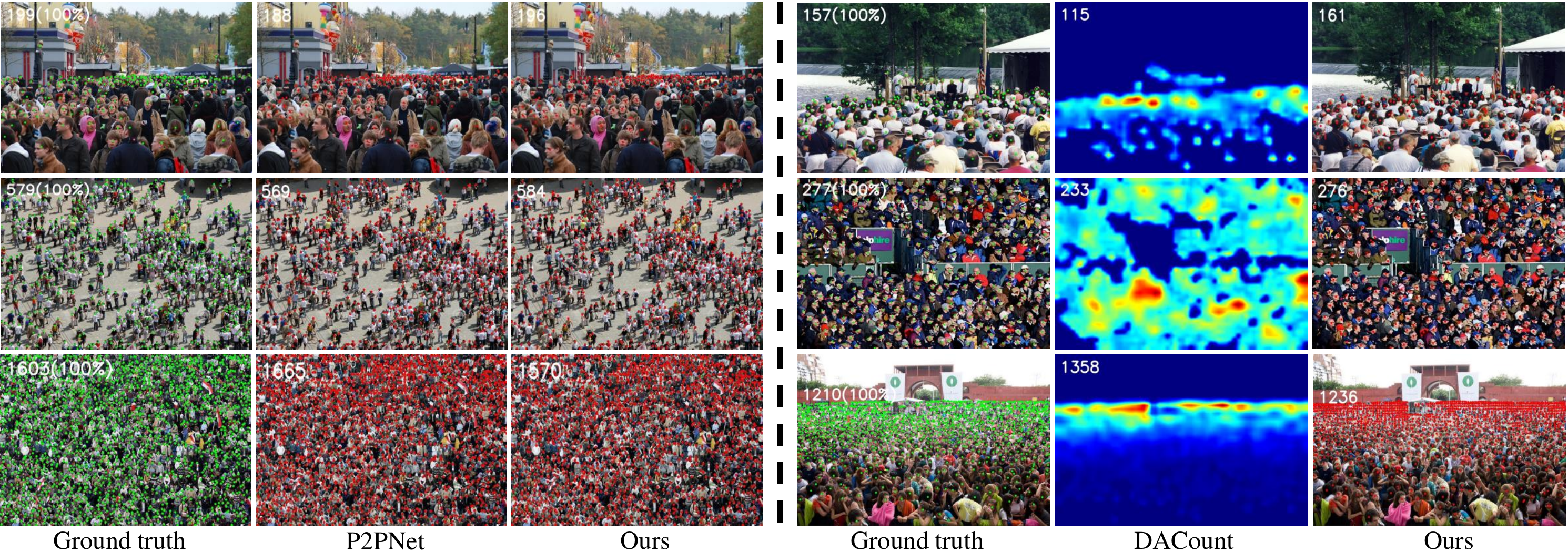}
  \vspace{-0.6em}
  \caption{Visualization of crowd counting prediction. On the left, it illustrates that our method is better than the baseline with 80\% annotations. On the right, it shows our method achieves better estimation than the semi-supervised DACount~\cite{lin2022semi} with 60\% annotations. As our method is point-based, it also gives the head location of people.  }
  \vspace{-1.5em}
  \label{fig:vis}
\end{figure*}

We compare our method with the state-of-the-art semi-supervised methods on SHA dataset, as shown in Table~\ref{tab:semi}. When 40\% training data are annotated, our method achieves the best performances by a large margin of 28.7, 27, 9, and 7.8 points, compared with MT~\cite{DBLP:conf/nips/TarvainenV17}, L2R~\cite{liu2018leveraging}, SUA~\cite{meng2021spatial}, and the recent DACount~\cite{lin2022semi}, respectively. Besides, when we set the ratio as 40\% and 30\%, our method also beats the DACount 4.3 and 3.1 points in terms of MAE. 


\begin{table}
    \begin{tabular}{l p{0.54cm}p{0.54cm}p{0.54cm}p{0.54cm}p{0.54cm}p{0.54cm}}
    \toprule
    \multicolumn{1}{c}{\multirow{2}{*}{Method}} & \multicolumn{2}{c}{90\%}      & \multicolumn{2}{c}{80\%}      & \multicolumn{2}{c}{50\%}       \\
    \cline{2-7} 
     & MAE           & MSE           & MAE           & MSE           & MAE           & MSE            \\
    \midrule
    PA~\cite{xu2021crowd}                                         & 67.4          & 103.8         & 67.7          & 103.7         & 70.5          & 105.0          \\
    \hline
    PA-PPM &52.5 &86.6 &54.7 &89.6 &62.7 &103.9\\
    SA-PPM(ours)                                        & \textbf{51.3} & \textbf{84.3} & \textbf{52.2} & \textbf{84.9} & \textbf{59.5} & \textbf{103.8} \\
    \bottomrule
    \end{tabular}
    \vspace{-1em}
  \caption{Comparison with Partial Annotation on SHA. Our method achieves performance gain consistently. }
  \vspace{-2em}
  \label{tab:part}
\end{table}


\subsection{Comparison with Partial Annotated Method}
Partial Annotated Crowd Counting~\cite{xu2021crowd} is a crowd counting setting which chooses to annotate one patch for each image in training set. In Table~\ref{tab:part}, we compare our performance with it. Quantitative results show that our method gains 16.1, 15.5, and 11 points improvement on three different annotation ratios with 90\%, 80\%, and 50\%, respectively. To ensure a fair comparison, we also train our framework under the setting of \cite{xu2021crowd}. And the results in Table~\ref{tab:part} prove that our proposed framework is also effective for partial annotation, achieving improvements of 14.9, 12 and 7.8 points under ratios of 90\%, 80\% and 50\% respectively. Additionally, Table \ref{tab:part} indicates that our sparse annotation, which captures diversity on both the image and patch levels, outperforms partial annotation under the same framework. 

Even though we get good performance on the annotation ratios in Table~\ref{tab:semi} and Table~\ref{tab:part}, our method cannot be implemented with the extremely annotated ratio, \eg 5\%. The reason is that the point-based approach relies on the number of training point samples. Too few samples lead to non-convergence of the neural network. It cannot replace the density map-based semi-supervised and partial annotated crowd counting approaches currently.

\subsection{Visualization}
On the left of Fig.~\ref{fig:vis}, we visualize the results of our model and the baseline with 80\% annotations. The baseline is improved significantly as we use the Performance Restoration Network to refine the point localization model. On the right of Fig.~\ref{fig:vis}, it shows the visualization results of our method and DACount~\cite{lin2022semi} with semi-supervision, both of which are with 50\% annotations. Our method achieves a better estimation of the count than DACount. Besides, different from the density map-based semi-supervised method which only estimates the number, our method also predicts the head localization of people as we take a point-based manner.  

\section{Conclusion}
Crowd counting is suffering from burdensome data annotation, which limits its application in real scenarios. To release the cost of annotation, we propose a novel Sparsely Annotated Crowd Counting setting, which explores the impact on the performance of the sparse annotations in each image. We also propose a simple yet effective framework to accomplish this setting, which consists of a Proposal Matching Network and a Performance Restoration Network. A basic point classifier is trained in Proposal Matching Network, which is used to generate pseudo point samples. Performance Restoration Network refines the classifier with the pseudo points. The extensive experiments on public crowd counting datasets verify the effectiveness of our setting and approach. 

Limitation: Our method cannot be implemented with the extremely low annotated ratio as too few point samples lead to non-convergence of the point-based PPM network. 
A two-stage manner is a potential way to overcome this limitation, in which the density map based method is used for the first stage and point-based method for the second stage. 

\small{\textbf{Acknowledgement.} This work was supported in part by the National Natural Science Foundation of China under Grant No. 62006182 and the Swiss National Science Foundation via the Sinergia grant CRSII5-180359.}

{\small
\bibliographystyle{ieee_fullname}
\bibliography{egbib}
}

\end{document}


\title{Appendix}

\author{First Author\\
Institution1\\
Institution1 address\\
{\tt\small firstauthor@i1.org}
\and
Second Author\\
Institution2\\
First line of institution2 address\\
{\tt\small secondauthor@i2.org}
}

\maketitle
\ificcvfinal\thispagestyle{empty}\fi


\section{Hard Threshold}

The predicted points of the PMN are noisy as the number of proposals $M$ is much larger than the number of annotated points $N^c$. Thus, it is critical to choose reliable pseudo points. 
Intuitively, we first propose a straightforward method to choose the positive point proposals with a hard threshold. For a point $\hat{p}_j$, the confidence score vector $\hat{c}_j$ is a 2D vector that represents the confidence of belonging to the background and foreground respectively. The hard threshold $\tau$ is used on the confidence score of the foreground to select pseudo points. For a point $\hat{p}_j=(x_j,y_j)$, its label is defined as:
\begin{equation}
L_{pseudo} = \left\{
\begin{aligned} 
1, &\quad \text{if} \quad \hat{c}_j > \tau \\
0, &\quad \text{otherwise}
\label{eq:hardthresh}
\end{aligned}\;.
\right.
\end{equation}
The pseudo point samples are selected by using this simple hard threshold on all of $M$ predicted points, which provides supervision for the PRN. 

Hard threshold can largely improve the performance of baseline, however, PPS which exclude inaccurate labels and explore the benefits of the sparse labels outperforms it.

\section{Annotate K or fewer people}
Annotating $K$ or fewer people(if the total number of an image is less than $K$) in an image can also be recognized as a kind of sparse annotation. We make a comparison between our annotation setting and K-annotation in Table~\ref{tab:K}.  Quantitative results show that our framework is also workable for K-annotation, but the performance of K-annotation is not as good as our annotation setting. The performance drops as a fixed $K$ can not provide patch-level diversity as sparse annotation. It is not enough to represent an extremely crowded image especially. 
\begin{table}[h]
\vspace{-0.8em}
\centering
\begin{tabular}{lp{0.34cm}p{0.34cm}p{0.34cm}p{0.34cm}p{0.34cm}p{0.34cm}p{0.34cm}p{0.34cm}}
\toprule
\multirow{2}{*}{Ratio} & \multicolumn{2}{c}{Baseline}  &    \multicolumn{2}{c}{SA(ours)}  &\multicolumn{2}{c}{$K$-annotation}  \\
\multicolumn{1}{c}{}                             & MAE   & MSE    & MAE           & MSE            & MAE   & MSE        \\
\midrule
80\%  ($K \approx 800$)  & 56.3          & 95.1           & \textbf{52.2}          & \textbf{84.9 }             &59.9  &101.2  \\
70\%  ($K \approx 600$)  & 65.1          & 112.4          & \textbf{54.5}          & \textbf{91.7}            &67.9  &126.4     \\
\bottomrule
\end{tabular}
\vspace{-1em}
\caption{The ablation of $K$-annotation on SHA.}
\label{tab:K}
\vspace{-1.2em}
\end{table}

\section{Performance of Crowd Localization}
Based on our PPM, our method is also capable of predicting crowd localization. 

In Table~\ref{tab:loc}, we make a comparison between our method and other Point Matching crowd localization methods. As Table~\ref{tab:loc} shows, our method achieves the best performances by a huge margin when $\sigma=4$. Moreover, when $\sigma=8$, we also obtain the best performance on Recall and F1-measure and only 0.6 lower than LPMN\cite{niu2022local} on Precision. 

\begin{table}[h]
  \centering
  \resizebox{\linewidth}{!}{
  \begin{tabular}{lccccccc}
    \toprule
    \multirow{2}{*}{methods} & \multirow{2}{*}{Ratio} & \multicolumn{3}{c}{$\sigma$ = 4}   & \multicolumn{3}{c}{$\sigma$ = 8} \\
    \multicolumn{1}{c}{}  & & Pre &Rec &F1-m&Pre &Rec &F1-m\\
    \midrule
    LPMN\cite{niu2022local} &100\% &-- &-- &-- &\textbf{80} &75 &76 \\
    CLTR~\cite{liang2022end} &100\%&43.6 &42.7 &43.2 &74.9 &73.5 &74.2 \\
    \hline
    PPM & 90\%   &\textbf{56.2} &\textbf{56.3} &\textbf{56.2} &79.4 &\textbf{79.5} &\textbf{79.5} \\
    PPM & 80\%   &54.4 &53.3 &53.8 &79.7 &77.9 &78.8\\
    \bottomrule
  \end{tabular}}
  \caption{Comparison of the localization performance on SHA.}
  \vspace{-1em}
  \label{tab:loc}
\end{table}

{\small
\bibliographystyle{ieee_fullname}
\bibliography{egbib}
}